\pdfoutput=1
\documentclass[11pt]{article}
\usepackage[]{acl}
\usepackage{times}
\usepackage{latexsym}
\usepackage{multirow}
\usepackage{graphicx}
\usepackage{booktabs}
\usepackage[T1]{fontenc}
\usepackage[utf8]{inputenc}
\usepackage{microtype}

\title{Experiments on Generalizability of BERTopic on Multi-Domain Short Text}

\author{Muriël de Groot\thanks{~Work done when student at the University of Amsterdam.} \\
  Albert Heijn, \\
  Ahold Delhaize, \\
  Amsterdam, The Netherlands \\
  \texttt{muriel.de.groot@ah.nl} \\\And
  Mohammad Aliannejadi \\
  IRLab, \\
  University of Amsterdam, \\
  Amsterdam, The Netherlands \\
  \texttt{~~m.aliannejadi@uva.nl} \\\And
  Marcel R. Haas \\
   Business Intelligence \\
  University of Amsterdam \\
  Amsterdam, The Netherlands \\
  \texttt{m.r.haas@uva.nl} \\}

\begin{document}
\maketitle
\begin{abstract}
Topic modeling is widely used for analytically evaluating large collections of textual data. One of the most popular topic techniques is Latent Dirichlet Allocation (LDA), which is flexible and adaptive, but not optimal for e.g. short texts from various domains. We explore how the state-of-the-art BERTopic algorithm performs on short multi-domain text and find that it generalizes better than LDA in terms of topic coherence and diversity.  We further analyze the performance of the HDBSCAN clustering algorithm utilized by BERTopic and find that it classifies a majority of the documents as outliers. This crucial, yet overseen problem excludes too many documents from further analysis. When we replace HDBSCAN with k-Means, we achieve similar performance, but without outliers. 

\end{abstract}

\section{Introduction}
\vspace*{-3mm}
Currently, one of the most widely used topic models in NLP is Latent Dirichlet Allocation (LDA) \citep{blei2003latent}. Although LDA is effective at modeling long conventional text collections such as news articles, it performs poorly on less conventional text such as short documents \citep{zuo2016word, hong2010empirical, zhao2011comparing, qiang2017topic}. Other algorithms like the biterm topic model (BTM) \citep{yan2013biterm} are successful in modeling topics in short texts. However, a common weakness between LDA and BTM is that the models cannot capture the context of words. As a result, the models lack generalizability across different domains, where different words are used to describe similar concepts \citep{zhao2014supervised, zhuang2010collaborative}.
BERTopic \citep{grootendorst2022bertopic} is a state-of-the-art topic modeling technique that could potentially overcome the mentioned limitations of LDA. BERTopic leverages pre-trained language embeddings and a class-based TF-IDF to create dense clusters, allowing for easily interpretable topics. 
We study and compare the performance of BERTopic and LDA on multi-domain short text corpora. Our goal is to compare how these models perform and generalize across multiple domains. \looseness=-1

To this aim, we conduct our experiments on a corpus of open-question responses on university course evaluations from various faculties, which ensures that the language used across the corpus is diverse.
Our experiments reveal that the HDBSCAN clustering utilized in BERTopic discards 74\% of the documents by labeling them as outliers, leading to results with little value in practice. Therefore, we propose using the k-Means clustering algorithm to maintain effectiveness while controlling the model's sensitivity to outliers. 

Our results show that BERTopic outperforms LDA in terms of different performance measures in various domains, confirming that it can generalize better across different domains on short text. 
Our contributions can be summarized as follows. (i) We apply BERTopic to unconventional data sets, characterized as short documents originating from different domains; (ii) we investigate the impact of replacing the original HDBSCAN clustering by k-Means clustering on the performance of BERTopic.

\begin{table*}
\footnotesize
    \vspace*{-4mm}
    \caption{Results. Ranging from 5 to 30 topics with steps of 5, topic coherence (TC) and topic diversity (TD) are shown for each topic model. All results are averaged across three runs. 
    Highest and lowest scores are marked in green and red, respectively.}
    \label{table:comparison new}
    \centering
    \vspace*{-3mm}
    \resizebox{\textwidth}{!}{
        \begin{tabular}{lcccccccc|c}
        \toprule
        & & \textbf{Faculty I} & \textbf{Faculty II} & \textbf{Faculty III} & \textbf{Faculty IV} & \textbf{Faculty V} & \textbf{All faculties} & \textbf{Average} & \textbf{20NG} \\
        \midrule \midrule
        \multirow{2}{*}{LDA} & TC & \textcolor{red}{-0.022} & -0.020 & -0.013 & 0.025 & -0.015 & 0.031 & -0.002 & 0.057 \\
        & TD &\textcolor{red}{0.358} & 0.528 & 0.414 & 0.621 & 0.500 & 0.718 & 0.523 & 0.752 \\
        \midrule
        \multirow{2}{*}{BERTopic HDBSCAN} & TC & 0.046 & 0.057 & 0.030 & 0.118 & \textcolor{teal}{0.145} & 0.091 & 0.081 & 0.166 \\
        & TD & 0.735 & 0.837 & 0.855 & 0.829 & \textcolor{teal}{0.903} & 0.880 & 0.840 & 0.902 \\
        \midrule
        \multirow{2}{*}{BERTopic k-Means} & TC & 0.028 & 0.033 & 0.028 & 0.050 & 0.029 & 0.033 & 0.032 & 0.113 \\
        & TD & 0.531 & 0.626 & 0.598 & 0.653 & 0.618 & 0.571 & 0.600 & 0.853 \\
        \bottomrule
        \end{tabular}
    }
    \vspace*{-4mm}
\end{table*}

\vspace*{-1mm}
\section{Experimental Methodology}
\vspace*{-3mm}
\paragraph{Metrics.} We compare the performance of LDA, BERTopic HDBSCAN, and BERTopic k-Means regarding topic coherence and diversity. Topic coherence is evaluated using normalized pointwise mutual information (NPMI) \citep{bouma2009normalized}. NPMI ranges from -1 to 1, where a higher NPMI indicates more strongly related words within a topic. Topic diversity is the percentage of unique words in the top-n words of all topics \citep{dieng2020topic} and is in the ranges [0,1]. Diversity near 0 indicates redundant topics, while values close to 1 indicate varied topics. \looseness=-1

We calculate these performance measures on the corpora of each faculty separately, as well as from all faculties combined. As such, we evaluate how scalable the LDA and BERTopic models are across domains with potentially varying vocabulary.

\paragraph{Data.} 
Our dataset consists of open-text comments of university students across five faculties, ranging from computer science to law students. Our dataset contains 62,522 raw responses. The length of the responses varies considerably among the faculties, ranging from 14 to 20 words median.
We also experiment on the 20NG dataset,\footnote{\url{http://qwone.com/~jason/20Newsgroups/}} which serves as a benchmark for validation. It comprises 11,096 news articles
across 20 categories. 
We further examine whether vocabulary and document length impact the performance differences between models in the main experiment. To assess the impact of document length, two samples are taken from the 20NG dataset. The sample that consists of short documents has a median document length of 16, which is equal to the median document length of the CER data set. In other words, this sample of short documents from the 20NG data set represents the same document length as the CER data set. The long document sample consists of documents containing between 60 and 100 words. \looseness=-1

\vspace*{-1mm}
\section{Results}
\vspace*{-3mm}
Table~\ref{table:comparison new} lists the results. BERTopic HDBSCAN can successfully create a university-wide topic model, with a topic coherence of 0.091 and a topic diversity of 0.880. In these metrics it outperforms LDA, which has coherence and diversity scores of 0.031 and 0.718, respectively. Moreover, differences in vocabulary among faculties were detected. To illustrate, students from Faculty I tend to use ’tutorial’, where students from faculty III use the word ’class’ or ’seminar’ to describe the same concept. Although vocabulary differences have been detected, there was no impact on the topic generation of BERTopic.
Furthermore, we observe that the topic coherence of BERTopic HDBSCAN declines when applied to shorter documents, making the model as susceptible to document length as LDA, see Table \ref{table:doc_len}. BERTopic k-Means is the least susceptible to short documents. \looseness=-1

\begin{table}
\footnotesize
    \vspace*{-2mm}
    \caption{Results of the document length experiment on the 20NG corpus. The performance difference of the models between long and short documents is denoted with $\delta$.}
    \label{table:doc_len}
    \centering
    \vspace*{-3mm}
    \begin{tabular}{ccccc}
    \toprule
    & & \textbf{Long} & \textbf{Short} & $\delta$ \\
    & & \textbf{documents} & \textbf{documents} & \\
    \midrule \midrule
    \multirow{2}{*}{LDA} & TC & -0.020 & -0.180 & -0.160 \\
    & TD & 0.614 & 0.656 & 0.042 \\
    \midrule
    BERTopic & TC & 0.080 & -0.065 & -0.145  \\
    HDBSCAN & TD & 0.910 & 0.972 & 0.062  \\
    \midrule
    BERTopic & TC & 0.058 & -0.023 & -0.081 \\
    k-Means & TD & 0.851 & 0.850 & -0.001 \\
    \bottomrule
    \end{tabular}
    \vspace*{-4mm}
\end{table}

A limitation of BERTopic HDBSCAN is the outlier generation. Approximately 74\% of the student responses are classified as outliers, rendering HDBSCAN inappropriate for the analysis of university course evaluations, in which every response matters. In contrast, k-Means does not generate outliers, creates interpretable topics, and is more robust to short documents than HDBSCAN and LDA. However, in terms of generalizability to a university-wide corpus, BERTopic k-Means performs less effectively, compared to BERTopic HDBSCAN, with a coherence of 0.032 and a diversity of 0.571. \looseness=-1

\vspace*{-2mm}
\section{Conclusions}
\vspace*{-3mm}
This paper explores the state-of-the-art BERTopic algorithm on open-question responses from course evaluations, with the goal of creating a topic model architecture that is scalable to a university-wide level. Besides the algorithm's default clustering algorithm, HDBSCAN, we investigated k-Means. We compared the models to a baseline LDA model. Although BERTopic HDBSCAN shows superior performance with regards to topic coherence and diversity in a university-wide model, it labels the majority of the documents as outliers. We propose to replace HDBSCAN with k-Means, because that algorithm does not generate outliers and is more robust to short documents. The absence of an all-encompassing topic model performance measure that takes outlier generation into account is a limitation in this research and should be considered in future work. \looseness=-1

\bibliography{custom}

\begin{thebibliography}{11}
\expandafter\ifx\csname natexlab\endcsname\relax\def\natexlab#1{#1}\fi

\bibitem[{Blei et~al.(2003)Blei, Ng, and Jordan}]{blei2003latent}
David~M Blei, Andrew~Y Ng, and Michael~I Jordan. 2003.
\newblock Latent dirichlet allocation.
\newblock \emph{Journal of machine Learning research}, 3(Jan):993--1022.

\bibitem[{Bouma(2009)}]{bouma2009normalized}
Gerlof Bouma. 2009.
\newblock Normalized (pointwise) mutual information in collocation extraction.
\newblock \emph{Proceedings of GSCL}, 30:31--40.

\bibitem[{Dieng et~al.(2020)Dieng, Ruiz, and Blei}]{dieng2020topic}
Adji~B Dieng, Francisco~JR Ruiz, and David~M Blei. 2020.
\newblock Topic modeling in embedding spaces.
\newblock \emph{Transactions of the Association for Computational Linguistics},
  8:439--453.

\bibitem[{Grootendorst(2022)}]{grootendorst2022bertopic}
Maarten Grootendorst. 2022.
\newblock Bertopic: Neural topic modeling with a class-based tf-idf procedure.
\newblock \emph{arXiv preprint arXiv:2203.05794}.

\bibitem[{Hong and Davison(2010)}]{hong2010empirical}
Liangjie Hong and Brian~D Davison. 2010.
\newblock Empirical study of topic modeling in twitter.
\newblock In \emph{Proceedings of the first workshop on social media
  analytics}, pages 80--88.

\bibitem[{Qiang et~al.(2017)Qiang, Chen, Wang, and Wu}]{qiang2017topic}
Jipeng Qiang, Ping Chen, Tong Wang, and Xindong Wu. 2017.
\newblock Topic modeling over short texts by incorporating word embeddings.
\newblock In \emph{Pacific-Asia Conference on Knowledge Discovery and Data
  Mining}, pages 363--374. Springer.

\bibitem[{Yan et~al.(2013)Yan, Guo, Lan, and Cheng}]{yan2013biterm}
Xiaohui Yan, Jiafeng Guo, Yanyan Lan, and Xueqi Cheng. 2013.
\newblock A biterm topic model for short texts.
\newblock In \emph{Proceedings of the 22nd international conference on World
  Wide Web}, pages 1445--1456.

\bibitem[{Zhao and Mao(2014)}]{zhao2014supervised}
Rui Zhao and Kezhi Mao. 2014.
\newblock Supervised adaptive-transfer plsa for cross-domain text
  classification.
\newblock In \emph{2014 IEEE International Conference on Data Mining Workshop},
  pages 259--266. IEEE.

\bibitem[{Zhao et~al.(2011)Zhao, Jiang, Weng, He, Lim, Yan, and
  Li}]{zhao2011comparing}
Wayne~Xin Zhao, Jing Jiang, Jianshu Weng, Jing He, Ee-Peng Lim, Hongfei Yan,
  and Xiaoming Li. 2011.
\newblock Comparing twitter and traditional media using topic models.
\newblock In \emph{European conference on information retrieval}, pages
  338--349. Springer.

\bibitem[{Zhuang et~al.(2010)Zhuang, Luo, Shen, He, Xiong, Shi, and
  Xiong}]{zhuang2010collaborative}
Fuzhen Zhuang, Ping Luo, Zhiyong Shen, Qing He, Yuhong Xiong, Zhongzhi Shi, and
  Hui Xiong. 2010.
\newblock Collaborative dual-plsa: mining distinction and commonality across
  multiple domains for text classification.
\newblock In \emph{Proceedings of the 19th ACM international conference on
  Information and knowledge management}, pages 359--368.

\bibitem[{Zuo et~al.(2016)Zuo, Zhao, and Xu}]{zuo2016word}
Yuan Zuo, Jichang Zhao, and Ke~Xu. 2016.
\newblock Word network topic model: a simple but general solution for short and
  imbalanced texts.
\newblock \emph{Knowledge and Information Systems}, 48(2):379--398.

\end{thebibliography}

\end{document}